\newcommand{\TypeOfDoc}{conf} 
\documentclass[letterpaper, 10 pt, conference]{IEEEtran} 
\pdfoutput=1

\usepackage[numbers]{natbib}
\usepackage{multicol}
\usepackage[bookmarks=true]{hyperref}
\usepackage{amsmath}
\usepackage{amssymb}
\usepackage{hyperref}
\usepackage{xspace}
\usepackage{graphicx}
\usepackage{subcaption} 
\usepackage{epstopdf}
\usepackage{listings}
\usepackage{ifthen}
\usepackage{bbm}
\usepackage{bm}
\usepackage{lipsum}
\usepackage{color} 
\usepackage{tikz}
\usepackage[capitalise]{cleveref}

\newcommand{\Cat}{\mathit{Cat}}
\newcommand{\Dir}{\mathit{Dir}}
\newcommand{\Ga}{\mathit{Ga}}

\DeclareMathOperator*{\argmax}{argmax}

\newcommand{\ifboeing}[1]{\ifthenelse{\equal{\TypeOfDoc}{boeing}}{\color{black}#1\color{black}\xspace}{}} 
\newcommand{\ifconf}[1]{\ifthenelse{\equal{\TypeOfDoc}{conf}}{\color{black}#1\color{black}\xspace}{}} 

\hypersetup{%
	pdfborder = {0 0 0}
}

\begin{document}
	\title{Hierarchical Bayesian Noise Inference \\for Robust Real-time Probabilistic\\ Object Classification}
	\author{Shayegan Omidshafiei, Brett T. Lopez, Jonathan P. How, John Vian}
	\maketitle
	
	\begin{abstract}
		Robust environment perception is essential for decision-making on robots operating in complex domains. Principled treatment of uncertainty sources in a robot's observation model is necessary for accurate mapping and object detection. This is important not only for low-level observations (e.g., accelerometer data), but for high-level observations such as semantic object labels as well. This paper presents an approach for filtering sequences of object classification probabilities using online modeling of the noise characteristics of the classifier outputs. A hierarchical Bayesian approach is used to model per-class noise distributions, while simultaneously allowing sharing of high-level noise characteristics between classes. The proposed filtering scheme, called Hierarchical Bayesian Noise Inference (HBNI), is shown to outperform classification accuracy of existing methods. The paper also presents real-time filtered classification hardware experiments running fully onboard a moving quadrotor, where the proposed approach is demonstrated to work in a challenging domain where noise-agnostic filtering fails.
	\end{abstract}
	
	\IEEEpeerreviewmaketitle
	
	\section{Introduction}\label{sec:introduction}
	The availability of inexpensive, portable vision sensors, parallelizeable object detection and classification algorithms, and general purpose GPU-based computational architectures make real-time scene understanding an increasingly-viable goal in robotics \cite{conf/iscas/LeCunKF10}. Localization and classification of objects in a robot's vicinity are needed, for instance, to perform complex robot-object interaction tasks or semantic labeling of the environment. 
	Moreover, decision-making based on semantic observations of a robot's surroundings can benefit by considering the probabilistic nature of these observations, allowing principled treatment of the sources of domain uncertainty. 
	For lightweight robots (e.g., small aerial vehicles), real-time object classification using transportable, low-power, affordable sensors such as monocular cameras is highly desirable. A unique trait of robotic platforms is locomotion, allowing observations of an object or scene from a variety of viewpoints. This motivates the need for a real-time, sequential object classification framework which uses the history of observations made by the robot throughout its mission. In contrast to na{\"i}ve reliance on frame-by-frame classification, sequential object labeling offers increased robustness against sources of uncertainty such as camera noise, varying lighting conditions, and occlusion. 
	
	Deep Convolutional Neural Networks (CNNs) are the state-of-the-art models for image classification, currently achieving up to 94\% accuracy on the ImageNet Large Scale Visual Recognition Challenge (ILSVRC) \cite{ILSVRC15}. Although the majority of object classification work has been conducted on still images \cite{conf/cvpr/SzegedyLJSRAEVR15,krizhevsky2012imagenet,journals/pami/FarabetCNL13, journals/corr/SermanetEZMFL13}, recent promising works on object localization \cite{he15deepresidual,conf/cvpr/GirshickDDM14} and video classification \cite{conf/cvpr/KarpathyTSLSF14,conf/cvpr/NgHVVMT15} also rely on CNNs. Video classification approaches, in particular, combine motion and single-frame information to predict labels. Extending this to the robotics setting is non-trivial, especially in exploration and navigation tasks where observations of a certain region in space may be low in frame-rate or where classification of static scenes from varying viewpoints (but with no object motion) is desired. For instance, the robot may observe an object, label it as belonging to a certain class, leave and re-enter the scene, and re-label the object as a different class based on observations from a new viewpoint. The robot then needs to make a decision regarding the underlying object class based on its history of past classifications. The need for classification filtering is further motivated since in the online setting, retraining the underlying image classifier is often infeasible due to inherent lack of labeled data and high computational cost. 
	
	
	\begin{figure}[t]
		\begin{subfigure}[t]{0.23\textwidth}
			\centering
			\includegraphics[height=0.9\textwidth]{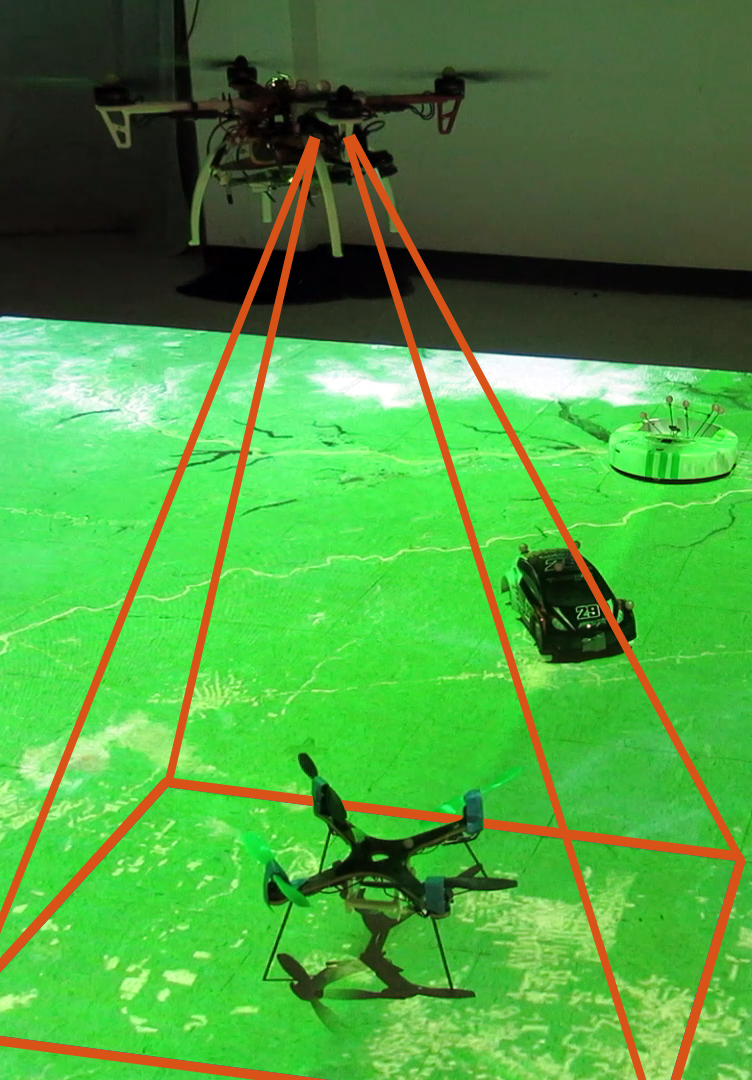}
			\caption{Classification robustness experiments conducted onboard moving quadrotor.}
			\label{fig:overview_quad_proj}
		\end{subfigure}
		\hfill
		\begin{subfigure}[t]{0.23\textwidth}
			\centering
			\includegraphics[height=0.9\textwidth]{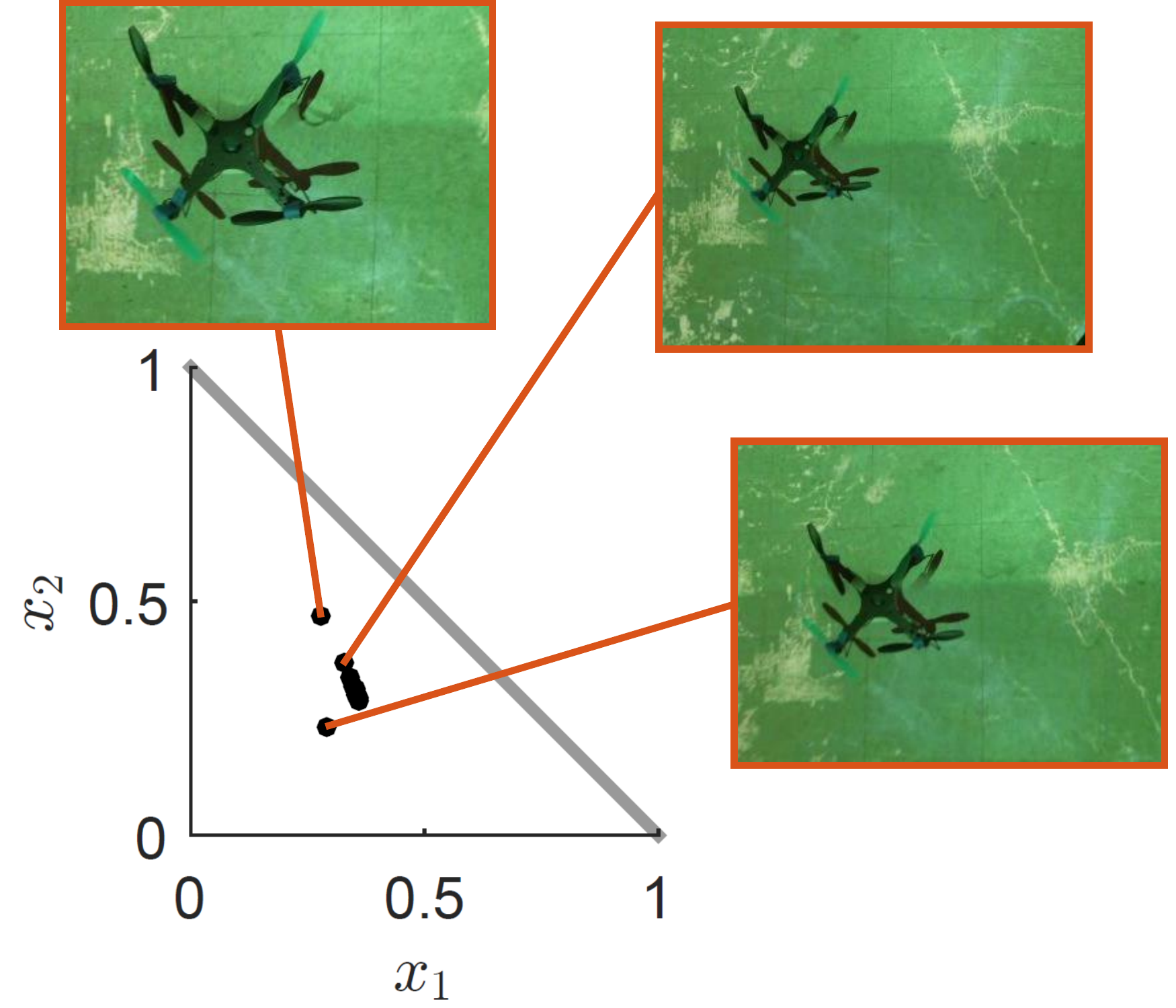}
			\caption{Example classification probability outputs on 2-simplex.}
			\label{fig:overview_fig_simplex}
		\end{subfigure}
		\caption{Real-time onboard probabilistic classification experiments in environments with varying lighting conditions, textures, motion blur, and other sources of uncertainty.}
		\label{fig:overview_fig}
	\end{figure}
	
	\begin{figure*}[t]
		\centering
		\begin{subfigure}[t]{0.19\textwidth}
			\centering
			\includegraphics[width=1\textwidth]{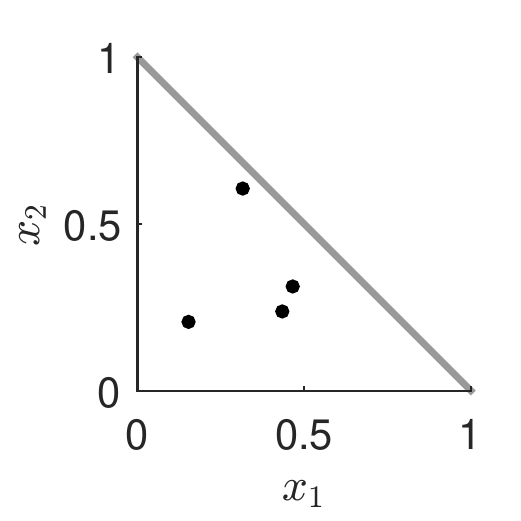}
			\caption{Classification probabilities of 4 images ($N=4$). Mean prediction probability is $(1/3,1/3,1/3)$.}
			\label{fig:obs_model_example}
		\end{subfigure}
		\hfill
		\begin{subfigure}[t]{0.19\textwidth}
			\centering
			\includegraphics[width=1\textwidth]{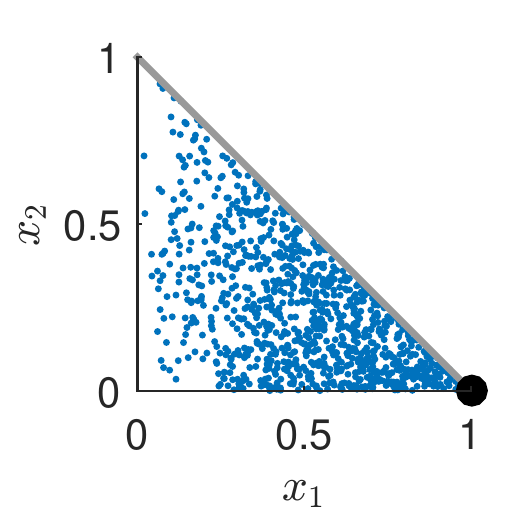}
			\caption{Noisy predictions for class $c=1$, $\theta_{c=1} = 1$.}
			\label{fig:obs_model_class_1}
		\end{subfigure}
		\hfill
		\begin{subfigure}[t]{0.19\textwidth}
			\centering
			\includegraphics[width=1\textwidth]{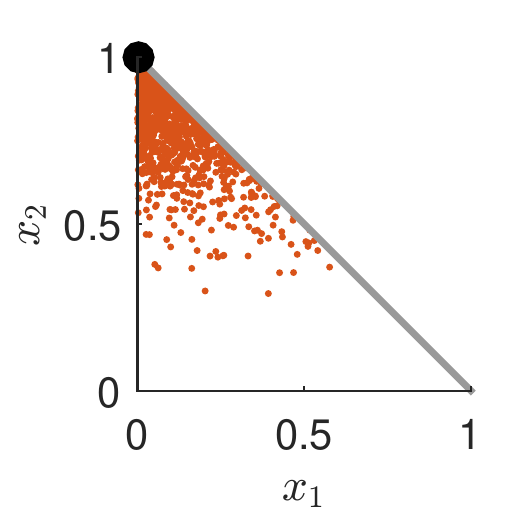}
			\caption{Noisy predictions for class $c=2$, $\theta_{c=2} = 6$.}
			\label{fig:obs_model_class_2}
		\end{subfigure}
		\hfill
		\begin{subfigure}[t]{0.19\textwidth}
			\centering
			\includegraphics[width=1\textwidth]{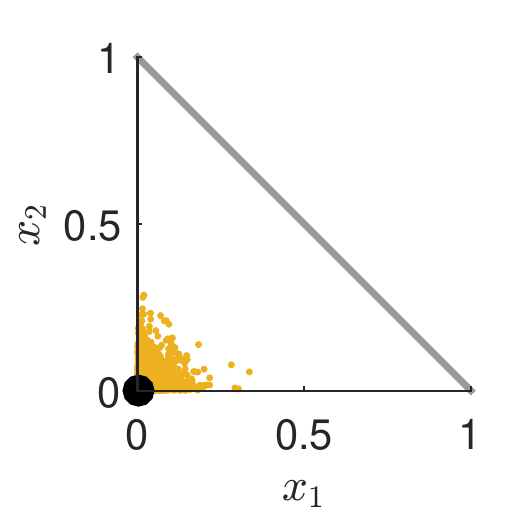}
			\caption{Noisy predictions for class $c=3$, $\theta_{c=3} = 20$.}
			\label{fig:obs_model_class_3}
		\end{subfigure}
		\hfill
		\begin{subfigure}[t]{0.19\textwidth}
			\centering
			\includegraphics[width=1\textwidth]{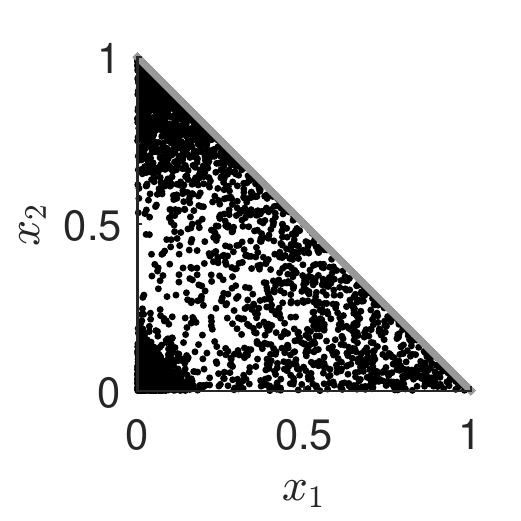}
			\caption{Unlabeled simplex observations require simultaneous inference of class labels and noise parameters.}
			\label{fig:obs_model_all_classes}
		\end{subfigure}
		\caption{Motivating example with 3 classes ($M=3$). The collection of predicted probabilities $x_{1:N}$ from a classifier may be uninformative (\cref{fig:obs_model_example}). Despite this, classifier noise modeling (\cref{fig:obs_model_class_1,fig:obs_model_class_2,fig:obs_model_class_3}) can lead to robust class inference. In this example, objects from class $c=1$ inherently have significantly higher prediction noise than classes $c=2$ and $c=3$. Therefore, observations $x_{1:N}$ (where $N=4$) in \cref{fig:obs_model_example} are more likely to come from an object with class $c=1$, despite having an uninformative mean probability of $(1/3,1/3,1/3)$.}
		\label{fig:example_observation_models}
	\end{figure*}
	
	
	In settings with high observation noise, or where training data is not representative of the data encountered during the mission, statistical analysis of the classifier's outputs can significantly improve underlying object classification, even without labeled data. As a motivating example, consider the 3-class scenario in \cref{fig:overview_fig_simplex}. The classifier predicts the probability of an input image (or feature descriptor) belonging to each class. A sequence of images (e.g., from varying viewpoints) results in a corresponding sequence of observed classification probabilities. \cref{fig:obs_model_example} illustrates such an observation sequence for 4 input images. For a given underlying object class, due to the aforementioned sources of uncertainty (or discrepancies between test and training data), the predicted outputs from the classifier are often noisy (\cref{fig:obs_model_class_1,fig:obs_model_class_2,fig:obs_model_class_3}). This prediction noise can make inference of the true underlying class (given the sequence of observations) nontrivial. For instance, the mean of the probability observations in \cref{fig:obs_model_example} is $(1/3,1/3,1/3)$, rendering the na{\"i}ve approach of using the average probability across all observations uninformative. 
	
	This paper's primary contribution is a hierarchical Bayesian framework for robust online object classification, using statistical noise modeling of classifier outputs to successfully identify objects in settings where noise-agnostic methods are demonstrated to fail. The paper also presents hardware results for real-time filtered object classification on a moving quadrotor, allowing accurate inference in noisy, real-world settings where training data is not representative of validation data. The entire processing pipeline is executed onboard the quadrotor, with classification and filtering running at approximately 13 frames per second. Although the hierarchical noise model introduced here is motivated by the robot object classification problem, it is a generalized approach which can be used for classification filtering in a broad range of applications. 
	
	\section{The Classification Filtering Problem}\label{sec:classification_filtering_problem}
	This section introduces the sequential-observation classification problem and related classifier filtering and consensus approaches. Given input feature vector $I_i$ at time $i$, an M-class probabilistic classifier outputs a prediction vector $x_i= (x_{i,1},\cdots,x_{i,m},\cdots, x_{i,M})$, where $x_{i,m}$ is the probability of the $i$-th feature vector belonging to the $m$-th class. Thus, $x_i$ resides in the $(M-1)$-simplex, such that
	\begin{align}
		x_{i,m} \geq 0 \hspace{30pt} ||x_i||_1=1.
	\end{align}
	
	In object classification, $I_i$ may be an image or a feature representation thereof, and $x_{i,m}$ represents the probability of the object in the image belonging to the $m$-th class (e.g., dog, house, car). This probabilistic classification can be conducted over a sequence of $N$ images $I_{1:N}$, resulting in a stream of class probability observations $x_{1:N}$. Sources of uncertainty such as camera noise, varying lighting conditions, motion blur, or occlusion cause $x_i$ to be a noisy classification of the true underlying object class. Therefore, simply labeling the object as belonging to the class with maximal probability $\argmax_{m} (x_{i,m})$ can lead to highly sporadic outputs as the image sequence progresses. A filtering or consensus scheme using the history of classifications $x_{1:N}$ is desirable.
	
	Prior work on aggregation of \emph{multiple} classifiers' predictions can be extended to single-classifier multi-observation filtering. Fixed classifier combination rules offer simplicity in implementation and run-time, at the cost of sub-optimality. The max-of-mean approach \cite{journals/tsmc/XuKS92} is one such consensus scheme, where the $m$-th class combined posterior probability, $x'_{m}$, is the mean of all observed probabilities,
	\begin{align}
	x'_{m} = \frac{1}{N} \sum_{i=1}^{N}x_{i,m}\,\,\,\, \forall m \in \{1,\cdots,M\}.
	\end{align}
	The highest mean-probability class is then chosen for the posterior object label.
	
	Another strategy is voting-based consensus \cite{Florian:2002:MCC:1118693.1118697, conf/icc/AshfaqJKR10}, where the posterior class $c$ is the one with the highest number of votes from all individual prediction observations $x_i$. More explicitly,
	\begin{align}
	c = \argmax_{c' \in \{1,\cdots,M\}} \sum_{i}\delta(c',\argmax_{m \in \{1,\cdots,M\}} x_{i,m})
	\end{align}
	where $\delta(\cdot,\cdot)$ is the Kronecker delta function. 
	
	The max-of-mean and voting consensus approaches do not exploit the probabilistic nature of classifier outputs, $x_i$. A Bayes filter offers a more principled treatment of the problem. Binary Bayes filters are a popular approach for occupancy grid filtering \cite{thrun2001}, where repeated observations of a grid cell are filtered to determine occupancy probability (a special case of multi-class filtering with $M=2$ classes, `occupied' or `empty'). They have also been used for multiple-camera object detection (another $M=2$ domain, where the object is either present or absent) \cite{conf/icra/CoatesN10}. The binary Bayes filter can be extended to M-class classification in the following recursive form,	
	\begin{align}
	P(c = m|I_{1:N}) & \propto P(I_N|c = m,I_{1:N-1})P(c = m|I_{1:N-1})\label{eq:bayes_filter_1}\\
	&\propto P(I_N|c = m)P(c = m|I_{1:N-1})\label{eq:bayes_filter_2}\\
	&\propto \frac{P(c = m|I_N)}{P(c = m)}P(c = m|I_{1:N-1})\label{eq:bayes_filter_3}
	\end{align}
	where Bayes rule is applied in \cref{eq:bayes_filter_1}, followed by a Markovian observation assumption in \cref{eq:bayes_filter_2}, and another application of Bayes rule in \cref{eq:bayes_filter_3}. $P(c=m)$ is the prior class distribution and $P(c=m|I_N) = x_{N,m}$. Note that this variant of Bayes filtering assumes a static underlying object class, and is henceforth referred to as a Static State Bayes Filter (SSBF).
	
	Although SSBF allows probabilistic filtering of classifier outputs, it treats the probability of each observation $x_i$ as equal in its update. In other words, it takes equal amount of evidence for a class to ``cancel out" the evidence against it. In settings with heterogeneous classifier performance, this may not be the best approach. For instance, one type of object may be particularly difficult to classify from certain viewing angles or under varying lighting conditions, increasing its probability of misclassifications compared to other object types. In our motivating example, for instance, \cref{fig:obs_model_class_1} illustrates a class $c=1$ which is particularly difficult to classify, with a near-uniform distribution of $x_i$ throughout the simplex, in contrast to the fairly confident classifications of class $c=3$ (\cref{fig:obs_model_class_3}). For this example, given a series of observations near the center of the simplex, $x_i = (1/3,1/3,1/3)$, the intuitive expectation is that update weight on underlying class $c=1$ should be higher than $c=3$, since the classifier typically shows much lower confidence when classifying objects of type $c=1$.
	
	\section{Hierarchical Bayesian Noise Inference}
	
	\begin{figure}[t]
		\centering
		\begin{subfigure}[t]{0.21\textwidth}
			\centering
			\includegraphics[width=.95\textwidth]{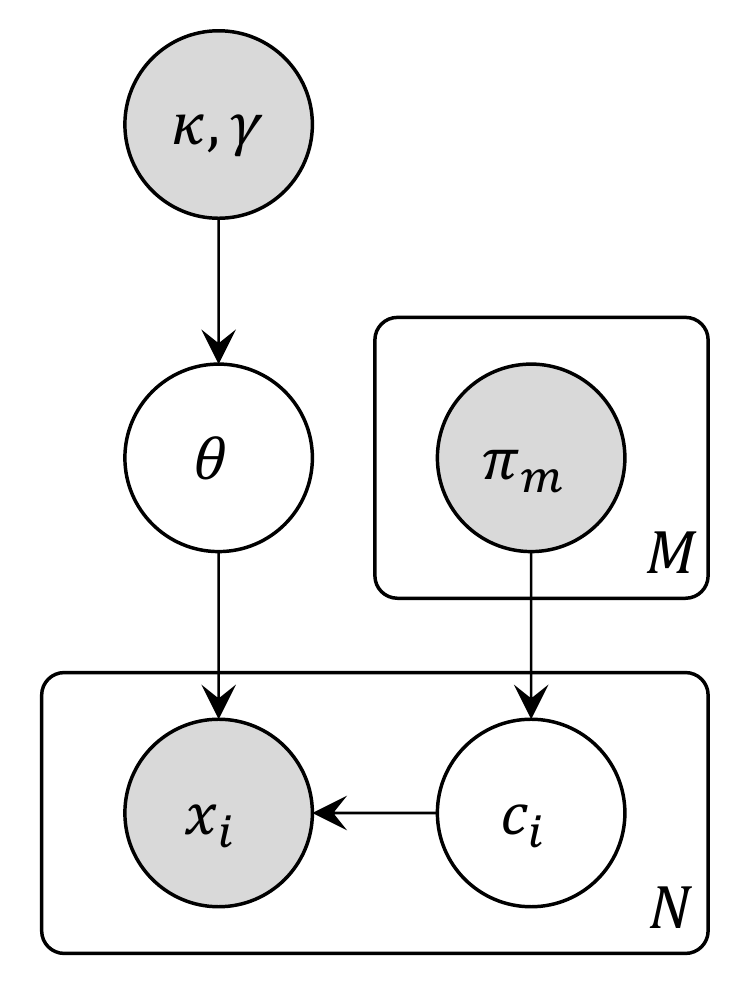}
			\caption{Simple Bayesian classification filter, with shared noise parameter $\theta$.}\label{fig:unique_obs_simple_bayes_plate_v2}
		\end{subfigure}
		\hfill
		\begin{subfigure}[t]{0.21\textwidth}
			\centering
			\includegraphics[width=0.95\textwidth]{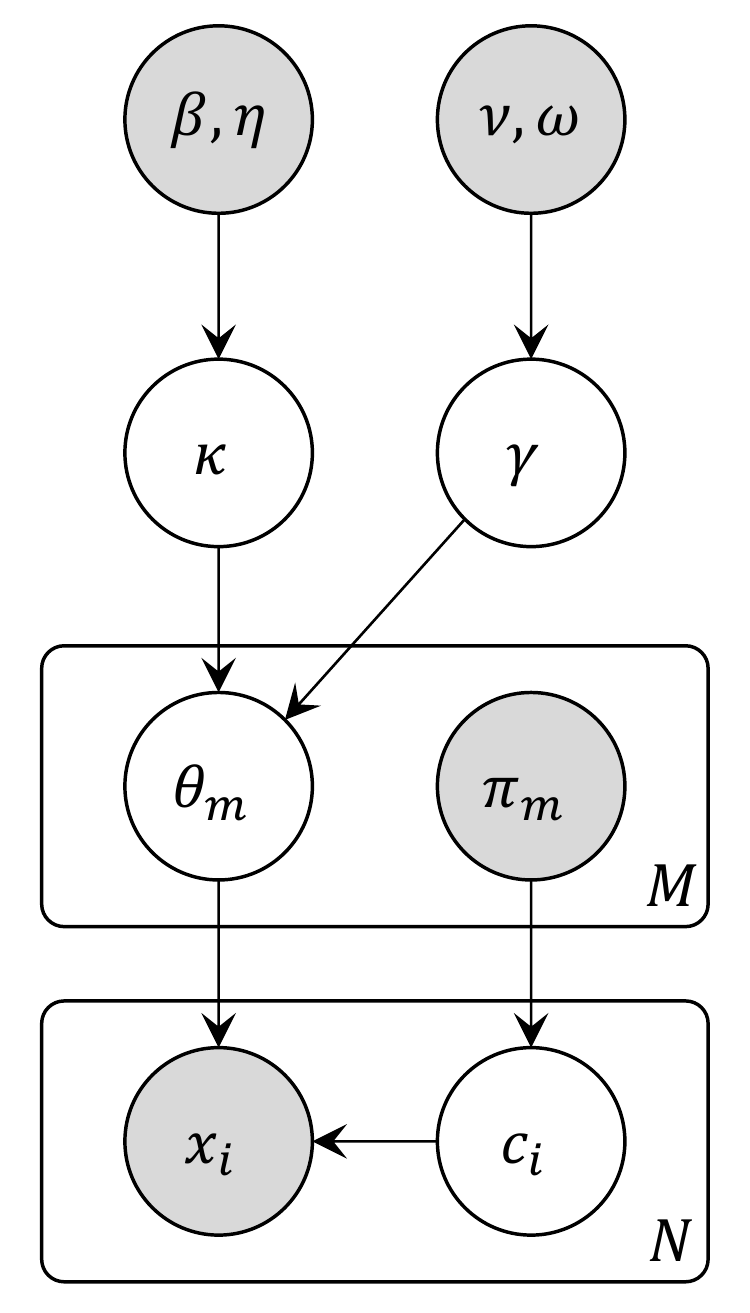}
			\caption{The HBNI model, with per-class noise parameters $\theta_m$ and shared hyperparameters, $\kappa, \gamma$.}\label{fig:unique_obs_hyperprior_bayes_plate_v2}
		\end{subfigure}
		\caption{Hierarchical Bayesian models for class inference.}
		\label{fig:bayes_plates}
	\end{figure}
	
	This section introduces Hierarchical Bayesian Noise Inference (HBNI), which models inherent heterogeneous classifier noise, allowing robust, probabilistic object classification in settings with limited observations $x_i$. HBNI is targeted towards online noise modeling, making this a 2-part inference problem. Given a collection of image class probability observations $x_{1:N} = \{x_1,\cdots,x_N\}$ (\cref{fig:obs_model_all_classes}), the true underlying class for each image $I_i$ must be inferred while simultaneously modeling the noise distribution associated with that class.
		
	Hierarchical Bayesian models \cite{good80, gelman_bayesian_2013} allow multi-level abstraction of uncertainty sources in a given problem. This is especially beneficial in settings where a layered set of uncertainty sources exist. In object classification, for instance, a shared parameterization of the classification noise across all classes may be modeled by a parameter $\theta$ (\cref{fig:unique_obs_simple_bayes_plate_v2}). A more flexible approach is to model per-class noise, using a set of parameters $\theta_{1:M} = \{\theta_1,\cdots,\theta_M\}$. Moreover, it may be beneficial to model the relationship between the noise parameters through a shared prior (\cref{fig:unique_obs_hyperprior_bayes_plate_v2}). This is useful in object classification, where sharing of high-level noise characteristics between classes is especially intuitive. Consider, for instance, a robot performing object classification using a malfunctioning camera, or in a domain with poor lighting. In this setting, observations of a single class may be significantly noisier than expected a priori, indicating the presence of a high-level source of uncertainty (e.g., a particularly noisy environment for imaging). This information should be shared amongst all class models, such that noise parameters can be updated to more accurately capture the uncertainty in the domain. Layered sharing of statistical information between related parameters is a strength of hierarchical Bayesian models, and has been demonstrated to increase robustness in posterior inference compared to non-hierarchical counterparts \cite{gelman_bayesian_2013, conf/icml/HugginsT15}.

	The graphical model of HBNI is illustrated in \cref{fig:unique_obs_hyperprior_bayes_plate_v2}. In this model, we assume a categorical prior on classes,
	\begin{align}
	c_{i} \sim \Cat(\pi_{1:M}),
	\end{align}
	where $\pi_{1:M}=\{\pi_1,\cdots,\pi_M\}$. A Dirichlet observation model is used for classifier outputs $x_i$, as they have support in the $(M-1)$-simplex,
	\begin{align}
	x_i \sim \Dir(\theta_{c_{i}}\vec{1}_{c_{i}}+\vec{1}),
	\end{align}
	where $\theta_{c_i} \geq 0$ is a scalar noise parameter for the associated class, $\vec{1}_{c_{i}}$ is an $M\times1$ categorical vector (with the $c_i$-th element equal to 1, remaining element equal to zero), and $\vec{1}$ is an $M\times1$ vector of ones. Therefore, each class observation $x_i$ has an associated class label $c_i$, which in turn links $x_i$ to the appropriate noise parameter $\theta_{c_i}$ (the $c_i$-th element of parameter set $\{\theta_1,\cdots,\theta_M\}$). This choice of parameterization offers two advantages. First, the selection of $\theta_{c_i}$ provides a direct, intuitive measure of noise for the classifier observations (as demonstrated in \cref{fig:obs_model_class_1,fig:obs_model_class_2,fig:obs_model_class_3}). $\theta_{c_i}$ is the concentration parameter of the Dirichlet distribution, and is related to the variance of the classification distribution. Low values of $\theta_{c_i}$ imply high levels of observation noise, and vice versa. A second advantage is that it simplifies the posterior probability calculations used within Markov chain Monte Carlo (MCMC) inference, as discussed below.
	
	A gamma prior is used for noise parameter $\theta_m$,
	\begin{align}
	\theta_m \sim \Ga(\kappa,\gamma),
	\end{align}
	where $\kappa$ and $\gamma$ themselves are treated as unknown hyperparameters. The role of $\kappa$ and $\gamma$ is to capture high-level sources of domain uncertainty, allowing sharing of cross-class noise statistics (as discussed earlier). Gamma priors (parameterized by ($\beta, \eta)$ and $(\nu,\omega)$) were also used for these hyperparameters in our experiments, although results showed low sensitivity to this prior choice.
	
	Thus, given a collection of image class probability observations $x_{1:N}$, the posterior probability of noise parameters and associated classes is,
	\begin{align}
		P&(\theta_{1:M},c_{1:N},\kappa,\gamma|x_{1:N}) \nonumber \\
		&\propto \prod_{i=1}^{N} P(x_i|\theta_{c_i}, c_i)P(c_i) \prod_{m=1}^{M}P(\theta_m|\kappa,\gamma)P(\kappa)P(\gamma)\\
		&= \prod_{i=1}^{N} \Big[\Dir(x_i;\theta_{c_i}\vec{1}_{c_i}+\vec{1})\Cat(c_i;\pi_{1:M})\Big]\nonumber\\ 
		& \phantom{{} =} \times \prod_{m=1}^{M}\Ga(\theta_m;\kappa,\gamma) \Ga(\kappa;\beta,\eta)\Ga(\gamma;\nu,\omega).\label{eq:theta_c_inference_orig}
	\end{align}
	
	\begin{figure}[t]
		\centering
		\hfill
		\includegraphics[width=0.5\textwidth]{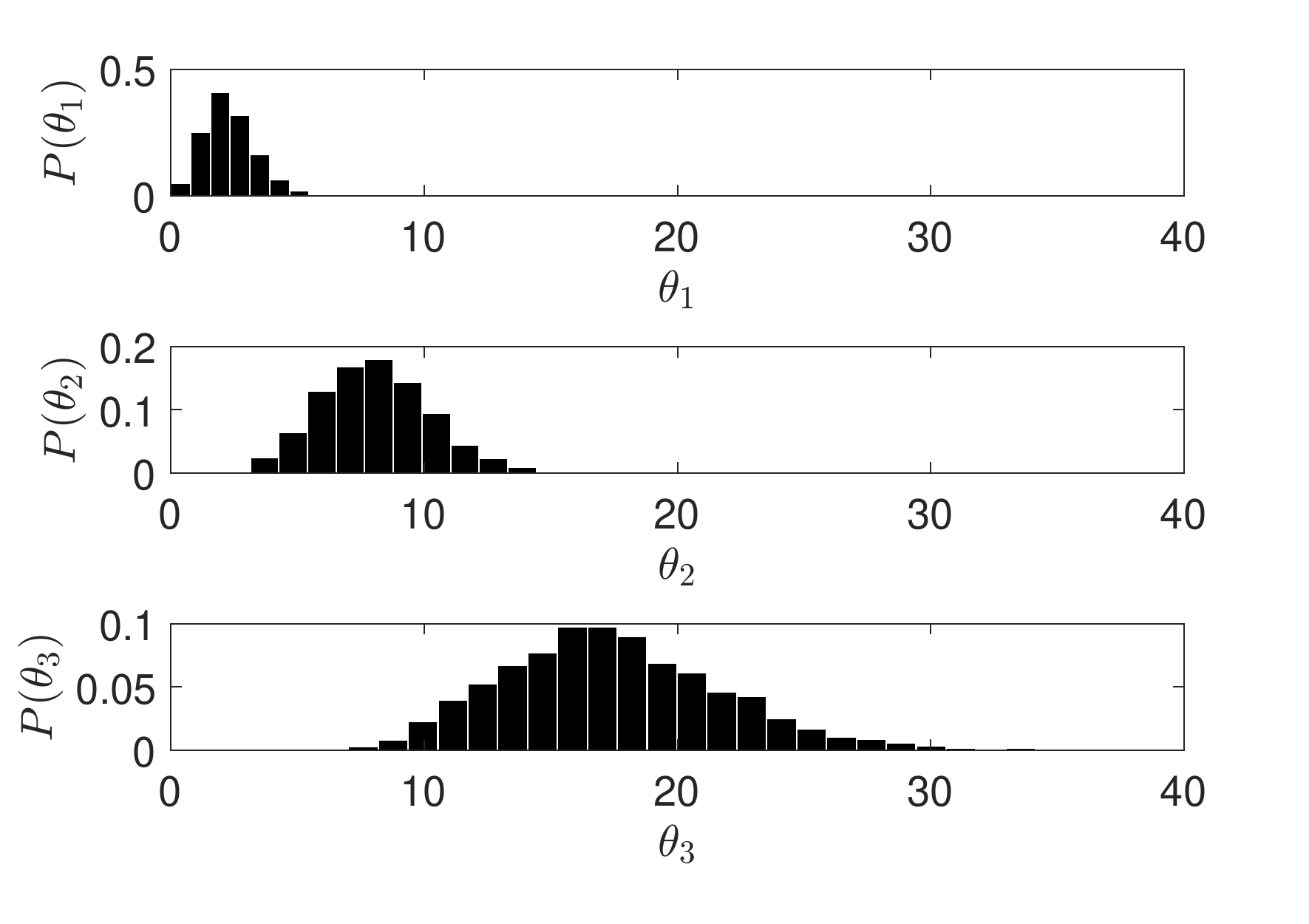}
		\caption{Inferred noise parameters $\theta$ for the $M=3$ classification problem illustrated in \cref{fig:example_observation_models}. True noise parameter values are $\theta_1=1$, $\theta_2=6$, $\theta_3=20$. }
		\label{fig:theta_posterior}
	\end{figure}
	
	\begin{figure}[t]
		\centering
		\begin{subfigure}[t]{0.22\textwidth}
			\centering
			\includegraphics[width=1\textwidth]{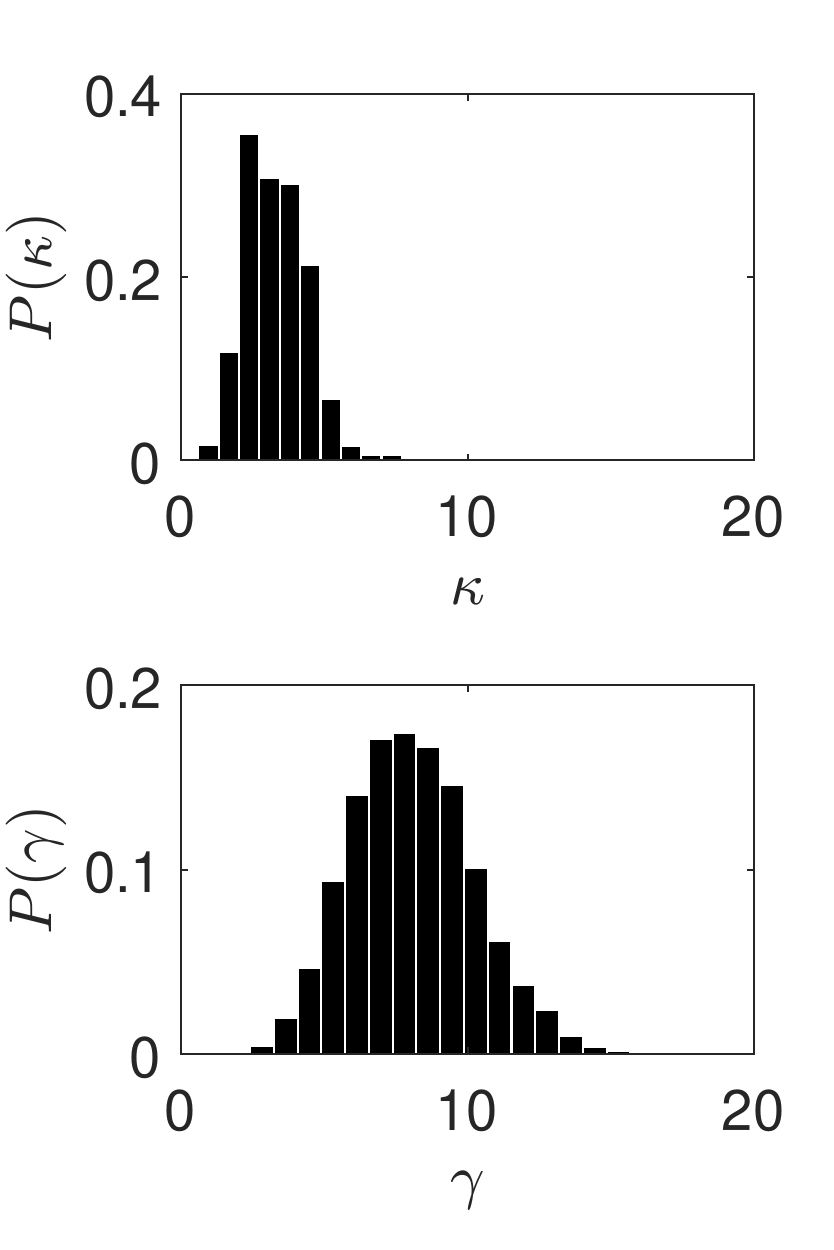}
			\caption{Posterior distributions of $\kappa$ and $\gamma$.}\label{fig:kg_posterior}
		\end{subfigure}
		\hfill
		\begin{subfigure}[t]{0.22\textwidth}
			\centering
			\includegraphics[width=1\textwidth]{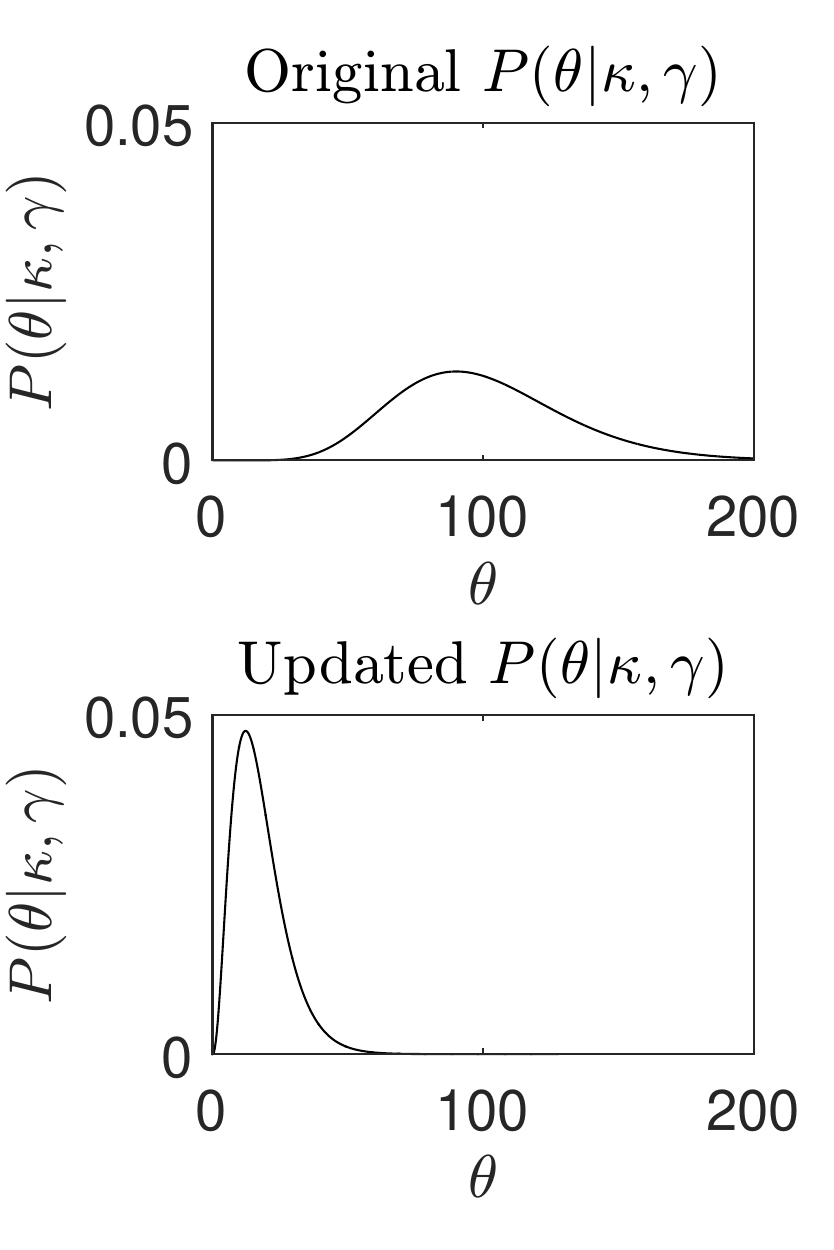}
			\caption{Change of $\theta$ distribution as a result of hyperparameter updates.}
			\label{fig:fhandle_theta_prior_posterior}
		\end{subfigure}
		\caption{Inference of high-level noise parameters $\kappa$, $\gamma$. Median hyperparameters were used for the plots on the right.}
		\label{fig:kg_theta_posteriors}
	\end{figure}
	
	This allows inference on underlying noise parameters $\theta_{1:M}$ and hyperparameters $\kappa$ and $\gamma$ based on the collection of observed data $x_{1:N}$, performed using an MCMC approach in this paper. Simplifications can be made to reduce the computational complexity of \cref{eq:theta_c_inference_orig}. Note, firstly, that the log of the categorical distribution is trivially computed,
	\begin{align}
		\log \Cat(c_i;\pi_{1:M}) = \log \pi_{c_i}.\label{eq:cat_log_final}
	\end{align}
	
	To efficiently compute $\log\Dir(x_i;\theta_{c_i}\vec{1}_{c_i}+\vec{1})$, consider a notation change. Letting $\bar{\alpha} = \{\alpha_1,\cdots,\alpha_M\}=\theta_{c_i}\vec{1}_{c_i}+\vec{1}$,
	\begin{align}
		\Dir(x_i;\bar{\alpha})=\frac{1}{B(\bar{\alpha})} \prod_{m=1}^{M}x_{i,m}^{\alpha_m-1},\label{eq:dirich_alpha_bar}
	\end{align}
	where $B(\cdot)$ is the Beta function. Based on the definition of $\bar{\alpha}$,
	\begin{align}
		\alpha_m-1=\begin{cases}
		\theta_{c_i}, & m=c_i.\\
		0, & \text{$m \neq c_i$}.
		\end{cases} \label{eq:alpha_cases}
	\end{align}
	Combining \cref{eq:alpha_cases} with \cref{eq:dirich_alpha_bar} and taking the log,
	\begin{align}
	\log\Dir&(x_i;\bar{\alpha})\nonumber\\
	&=-\log B(\bar{\alpha}) + \theta_{c_i}\log x_{i,c_i}\\
	&=-\sum_{m=1}^{M}\log\Gamma(\alpha_m)+\log\Gamma(\sum_{m=1}^{M}\alpha_m) + \theta_{c_i}\log x_{i,c_i},\label{eq:log_dir_beta_expanded}
	\end{align}
	where $\Gamma$ is the gamma function. Note that as per \cref{eq:alpha_cases}, 
	\begin{align}
		\log\Gamma(\alpha_m)=\begin{cases}
		\log\Gamma(1+\theta_{c_i}), & m=c_i.\\
		0, & \text{$m \neq c_i$}.
		\end{cases} \label{eq:log_gamma_cases}
	\end{align}
	and $\sum_m \alpha_m = M + \theta_{c_i}$. Thus, the log-posterior of the Dirichlet term can be computed as,
	\begin{align}
	\log\Dir&(x_i;\bar{\alpha})\nonumber\\	&=-\log\Gamma(1+\theta_{c_i})+\log\Gamma(M+\theta_{c_i}) +\theta_{c_i}\log x_{i,c_i}.\label{eq:log_dir_beta_final}
	\end{align}

	Finally, the log-probability of $\theta_m$ (and similarly $\kappa$, $\gamma$) can be computed,
	\begin{align}
	\log\Ga(\theta_m;\kappa,\gamma) &= \log \frac{\theta_m^{\kappa-1}\exp{(-\frac{\theta_m}{\gamma})}}{\gamma^{\kappa}\Gamma(\kappa)}\\
	&\propto (\kappa-1)\log\theta_m - \frac{\theta_m}{\gamma}\label{eq:gamma_log_final}
	\end{align}
	To summarize, the log of \cref{eq:theta_c_inference_orig} can now be computed by combining \cref{eq:cat_log_final,eq:log_dir_beta_final,eq:gamma_log_final}. An MCMC approach is used to calculate the associated posterior distribution over the noise parameters $\theta_{1:M}$ and hyperparameters $\kappa$ and $\gamma$. This then allows a collection of \emph{new} observations $x_{1:N}$ to be filtered using the posterior noise distributions, resulting in a probability of each possible class given the observation history,
	\begin{align}
		P&(c=m|x_{1:N},\theta_{1:M}) \nonumber\\
		&\propto P(x_{1:N}|\theta_m,c=m)P(c=m)\\
		&=P(x_N|\theta_m,c=m)\Big[\prod_{i=1}^{N-1}P(x_i|\theta_m,c=m)\pi_m\Big],\label{eq:classification_update_original}
	\end{align}
	where $c$ is conditionally independent of $\kappa_\theta$ and $\gamma_\theta$ given $\theta_{1:M}$, allowing us to drop the hyperparameter terms.
	
	Recall $P(x_N|\theta_m,c=m) = \Dir(x_N;\theta_{m}\vec{1}_{m}+\vec{1})$, the Dirichlet density at $x_N$. Let $(\psi_m)_{i} \triangleq P(c=m|x_{1:i},\theta_{1:M})$, such that \cref{eq:classification_update_original} leads to a compact recursive update rule for the class probability given inferred noise distribution $\theta_{1:M}$ and a new observation $x_N$,
	\begin{align}
	\begin{cases}
	 (\psi_m)_0 &= \pi_m.\\
	 (\psi_m)_N &=\Dir(x_N;\theta_{m}\vec{1}_{m}+\vec{1})(\psi_m)_{N-1}.
	\end{cases}\label{eq:recursive_HBNI}
 	\end{align}
	
	To summarize, the proposed HBNI approach uses a collection of \emph{unlabeled} classification observations $x_i$ to calculate a posterior distribution on noise parameters $\theta_{1:M}$ for each object class, and shared hyperparameters $\kappa$ and $\gamma$. These noise distributions are subsequently used to infer the (filtered) probability of each object class given a sequence of raw classification observations $x_{1:N}$.
	
	
	\section{Simulated Experiments}
	As stated earlier, an MCMC approach is used to compute the posterior over $\theta_{1:M}$, $\kappa$, and $\gamma$. Specifically, the experiments conducted use a Metropolis-Hastings (MH) \cite{metropolis53, hastings70} sampler with an asymmetric categorical proposal distribution for underlying classes $c_i$, with high weight on the previously-proposed class and low weight on remaining classes (with uniform random initialization). Gaussian MH proposals are used for transformed variables $\log(\theta_m)$, $\log(\kappa)$, and $\log(\gamma)$.
	
	\cref{fig:theta_posterior} shows noise parameter ($\theta_m$) posterior distributions for the $M=3$ problem outlined in \cref{fig:example_observation_models}. Parameter inference was conducted using only $N=15$ classification observations $x_i$ (5 from each class), with 8000 samples of a Metropolis-Hastings chain (6000 burn-in iterations and a thinning period of 10 between each sample). Despite the very limited number of observations, the posterior distributions provide reasonable inferences of the true underlying noise parameters. 
	
	Hyperparameter ($\kappa$, $\gamma$) posteriors are shown in \cref{fig:kg_posterior}. Recall that these parameters allow sharing of high-level noise information across multiple classes, and capture trends in outputs $x_i$ that indicate overall shifts in classification confidence levels due to domain-level uncertainty. To test sensitivity of $\theta_m$ inference to the hyperparameters, priors for $\kappa$ and $\gamma$ were chosen such that (on average) they would indicate very high values of $\theta_m$ (as seen in \cref{fig:fhandle_theta_prior_posterior}, top). This sets a prior expectation of near-perfect outputs from classifiers (median $\theta_m=100$). However, given only $N=15$ classifier output observations, the posteriors of $\kappa$ and $\gamma$ shift to indicate much lower overall classification certainty, or lower $\theta_m$ (\cref{fig:fhandle_theta_prior_posterior}, bottom). The prior distribution $P(\theta_m|\kappa,\gamma)$ has now shifted to better capture the range of noise parameters expected in the domain. This sharing of high-level noise statistics improves filtering of subsequent observations (even if from an entirely new class).
	
\begin{figure*}[t]
		\centering
		\begin{subfigure}[t]{0.16\textwidth}
			\centering
			\includegraphics[trim={0 20pt 0 20pt},clip, width=1\textwidth]{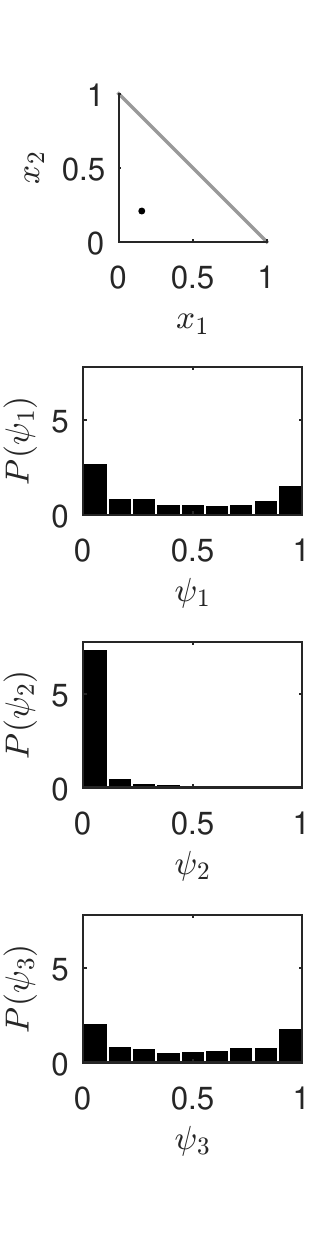}
			\caption{N=1}
			\label{fig:fhandle_class_posterior_num_obs_1}
		\end{subfigure}
		\hfill
		\begin{subfigure}[t]{0.16\textwidth}
			\centering
			\includegraphics[trim={0 20pt 0 20pt},clip, width=1\textwidth]{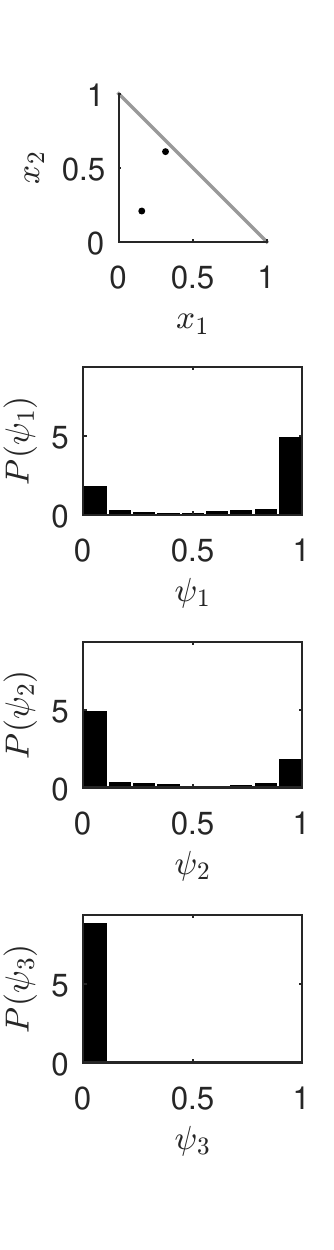}
			\caption{N=2}
			\label{fig:fhandle_class_posterior_num_obs_2}
		\end{subfigure}
		\hfill
		\begin{subfigure}[t]{0.16\textwidth}
			\centering
			\includegraphics[trim={0 20pt 0 20pt},clip, width=1\textwidth]{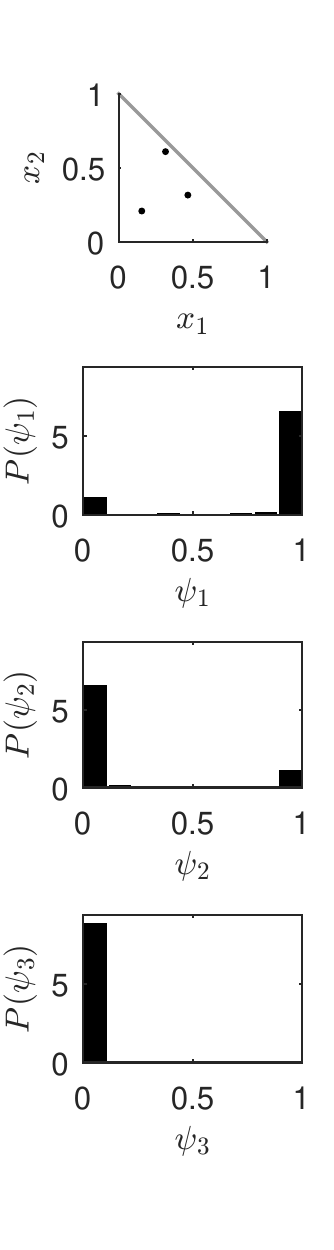}
			\caption{N=3}
			\label{fig:fhandle_class_posterior_num_obs_3}
		\end{subfigure}
		\hfill
		\begin{subfigure}[t]{0.16\textwidth}
			\centering
			\includegraphics[trim={0 20pt 0 20pt},clip, width=1\textwidth]{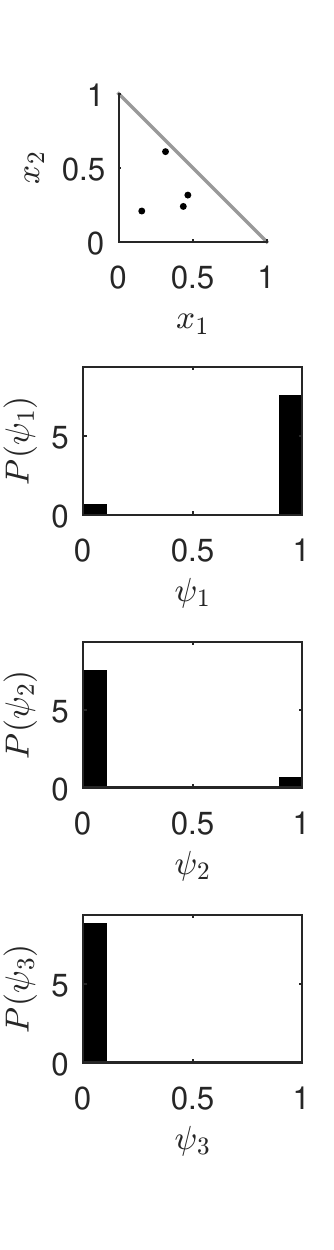}
			\caption{N=4}
			\label{fig:fhandle_class_posterior_num_obs_4}
		\end{subfigure}
		\hfill
		\begin{subfigure}[t]{0.075\textwidth}
			\centering
			\includegraphics[trim={0 20pt 0 20pt},clip, width=1\textwidth]{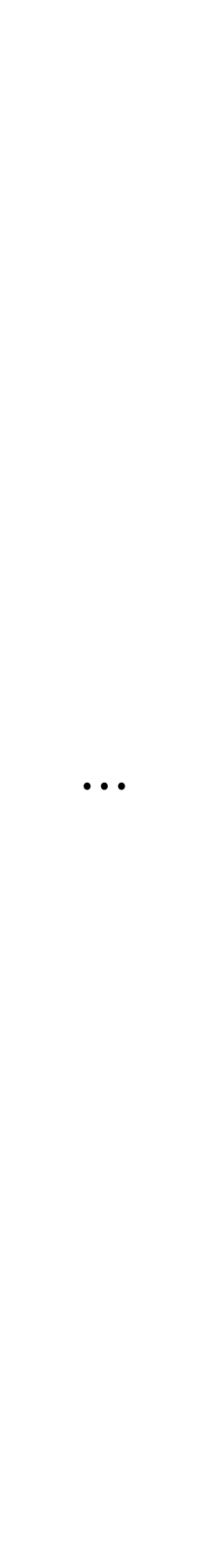}
		\end{subfigure}
		\hfill
		\begin{subfigure}[t]{0.16\textwidth}
			\centering
			\includegraphics[trim={0 20pt 0 20pt},clip, width=1\textwidth]{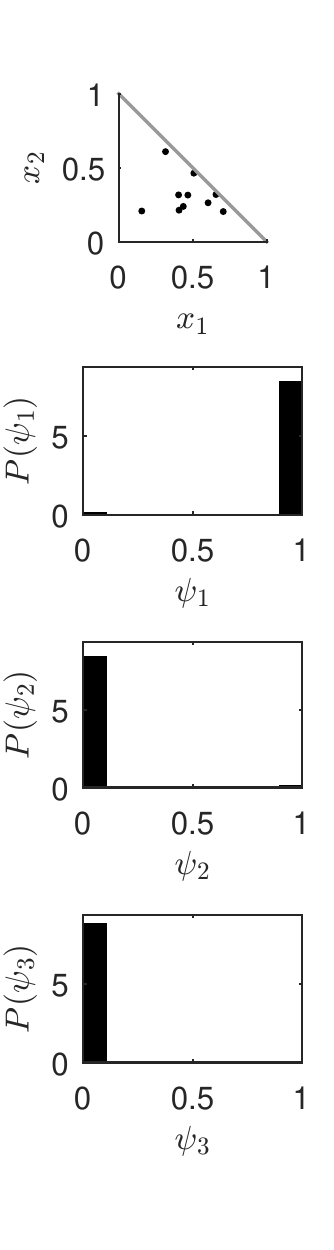}
			\caption{N=10}
			\label{fig:fhandle_class_posterior_num_obs_10}
		\end{subfigure}
		\caption{Sequential classification filtering for the example outlined in \cref{fig:example_observation_models}. In each column, the histograms show the `probability density of probabilities' for the classes, given observations $x_{1:N}$ of a single object. The true underlying class is $c=1$ in this example.}
		\label{fig:sequential_classification}
	\end{figure*}

	In \cref{fig:sequential_classification}, we consider the temporal behavior of filtered classifications using \cref{eq:recursive_HBNI}, specifically for the original motivating  $N=4$ observation problem introduced in \cref{sec:introduction} and \cref{fig:obs_model_example}. The true underlying class being observed in these experiments is $c=1$. Note that these results use the noise model inferred from the limited set of $N=15$ observations discussed above, and not the large number of observations indicated for illustration purposes in \cref{fig:obs_model_all_classes}. Since HBNI yields a posterior distribution on $\theta_m$ for each $m \in \{1,\cdots,M\}$, then given some new set of observations, $(\psi_m)_N = P(c=m|x_{1:N},\theta_{1:M})$ itself has a distribution with support $[0,1]$. More explicitly, \cref{fig:sequential_classification} plots the `probability of probabilities' for the various classes, given \emph{new} observations $x_{1:N}$ and previously-inferred noise distributions $\theta_m$. 
	
	With only a single observation (\cref{fig:fhandle_class_posterior_num_obs_1}), posterior class probability is split between $c=1$ (since it has the highest inferred observation noise) and $c=3$ (since the observation is near its corner point). As the number of observations increases (\cref{fig:fhandle_class_posterior_num_obs_2,fig:fhandle_class_posterior_num_obs_3}), so does confidence in the underlying class $c=1$. At $N=4$ (\cref{fig:fhandle_class_posterior_num_obs_4}), $c=1$ is the most likely class by a large margin, despite the mean of the observations being $(1/3,1/3,1/3)$. This matches the intuitive outcome discussed in \cref{sec:classification_filtering_problem}. HBNI allows modeling of the true underlying noise model (\cref{fig:example_observation_models}), leading to confident posterior estimates of underlying class labels even in cases with a limited number of observations. 
	
\begin{figure}[t]
		\centering
		\includegraphics[width=0.5\textwidth]{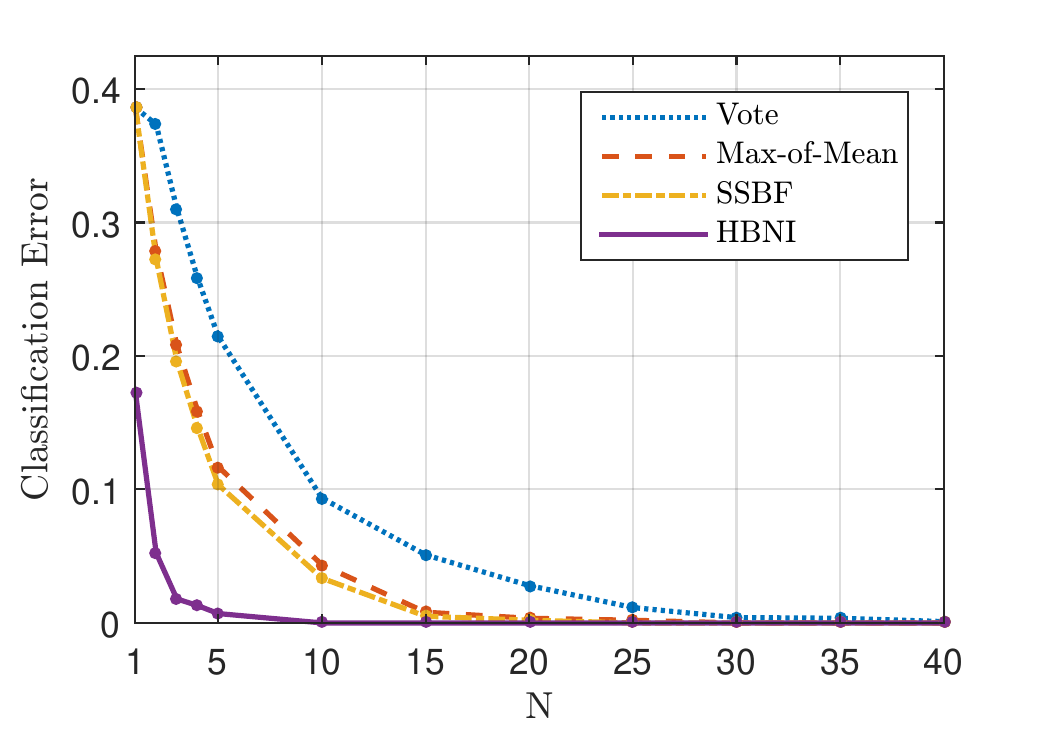}
		\caption{Performance comparison of classification filtering/consensus methods.}
		\label{fig:fhandle_method_comparison}
	\end{figure}

	Performance of HBNI, as measured by classification error, is evaluated against the voting, max-of-mean, and SSBF methods discussed in \cref{sec:classification_filtering_problem}. \cref{fig:fhandle_method_comparison} presents this comparison for varying number of classification observations, with 2000 trials used to calculate error for each $N$. As expected, the voting method performs poorly as it disregards the probabilistic nature of class observations $x_{1:N}$. Convergence trends for max-of-mean and SSBF are fairly similar, although SSBF performs slightly better and has the added benefit of probabilistic class posteriors. HBNI significantly outperforms the other methods, requiring only 10 observations to converge to the true object class for all trials. In general, the other methods need 4-5 times the number of observations to match HBNI's performance. One interesting result is that for $N=1$, class predictions for voting, max-of-mean, and SSBF are equivalent (as are classification errors). However, due to noise modeling, HBNI can make a more informed decision regarding underlying class even with 1 observation, leading to lower classification error. 
	
	\section{Hardware Experiments} 
	We now consider classification filtering on a hardware platform in order to ascertain whether noise modeling is beneficial in real-world settings. This section outlines real-time, probabilistic classification filtering experiments conducted onboard a moving quadrotor. 
	
	\subsection{Classification Framework and Training}
	Object classifier training is conducted on a dataset consisting of 3 target vehicle classes (`iRobot', `Quadrotor', `Racecar') in a well-lit room, using a Logitech C615 webcam at QVGA resolution (examples in \cref{fig:training_images_overview}). 100 snapshots of each object type are used for training, with crops and mirror images included to increase the classifier's translational and rotational invariance. Image feature extraction is done using a CNN-based architecture implemented in the open source Caffe framework \cite{jia2014caffe}. Input images are center-cropped with 10\% padding (to increase the classifier's robustness against non-focal objects near the image border) and resized to 227$\times$227 (input resolution of the CNN used). Features are extracted from the \texttt{fc8} (fully connected layer 8) of an AlexNet \cite{krizhevsky2012imagenet} trained on the ILSVRC2012 dataset \cite{ILSVRC15}. The resulting features are used to train a set of SVMs, with a one-vs-one approach used for multi-class classification (a set of $M(M-1)/2$ binary classifiers making decisions over each possible pairing of the $M$ classes). This offers higher modularity than a one-vs-all approach, as any of the binary classifiers can be removed if their associated objects of interest are not expected to appear during the mission. Additionally, the one-vs-one approach has significantly shorter training time than the one-vs-all approach, with negligible performance decrease \cite{milgram:inria-00103955}. Since SVMs are inherently discriminative classifiers, class probabilities $x_i$ for a given image $I_i$ are calculated using Platt Scaling \cite{Platt99probabilisticoutputs}, by fitting a sigmoid function to the SVM classification scores.
	
	To summarize, the classification pipeline accepts an input image $I_i$ at each timestep $i$, center-crops and resizes it, extracts features using the \texttt{fc8} layer of the CNN, performs multi-class probabilistic classification using a set of SVMs with Platt Scaling, and finally outputs vector $x_i$ indicating the probability of each object class for the given image. Probability outputs are then accumulated for sequential classification filtering comparisons. 

	\subsection{Hardware Setup}
\begin{figure}[t]
		\centering
		\includegraphics[width=0.49\textwidth]{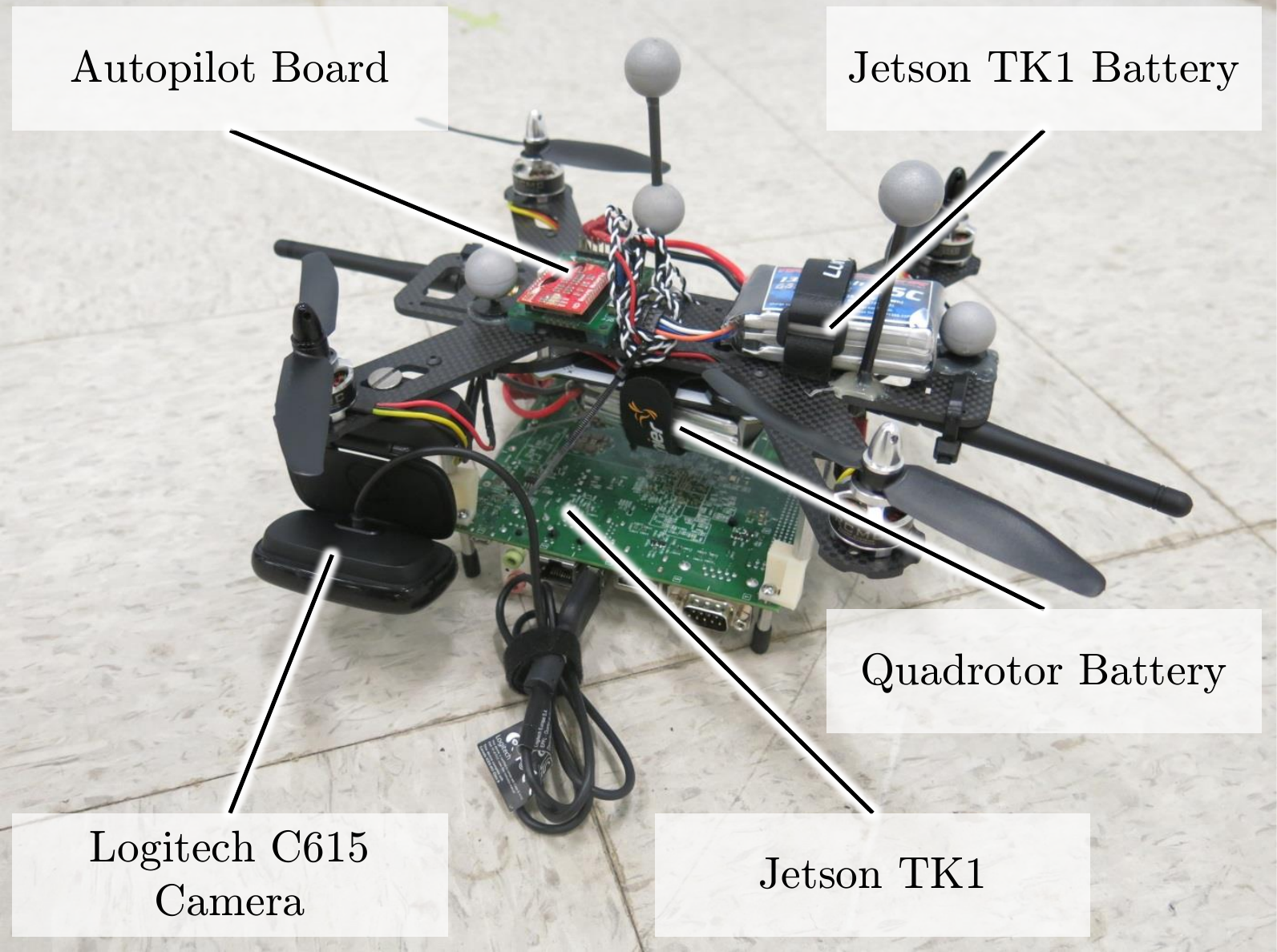}
		\caption[]{Hardware overview. Luminier QAV250 quadrotor frame with in-house designed autopilot board, Logitech C615 camera, onboard NVIDIA Jetson TK1 with dedicated battery for real-time object classification filtering.}
		\label{fig:jetson_quad_overview}
	\end{figure}

	A Luminier QAV250 quadrotor with a custom autopilot is used for the majority of experiments (\cref{fig:jetson_quad_overview}). A DJI F450 frame is used for some experiments (\cref{fig:overview_quad_proj}), as it can carry a larger capacity battery and has longer flight time. A Logitech C615 webcam is installed on one of the quadrotor arms for image capture. Image classification and HBNI filtering are executed on an onboard NVIDIA Jetson TK1 with 192 CUDA cores. A dedicated 3-cell 2100mAh lithium-ion polymer battery is used to power the Jetson, with typical operation durations between 2 to 4 hours (varying with computational load). Runtime for the Caffe-based classifier is 73$\pm$6ms per frame, and the entire pipeline (including communication and filtering) executes fully onboard at approximately 13 frames per second.
		
	\subsection{Results}
\begin{figure}[t]
		\centering
		\begin{subfigure}[t]{0.115\textwidth}
			\centering
			\includegraphics[width=1\textwidth]{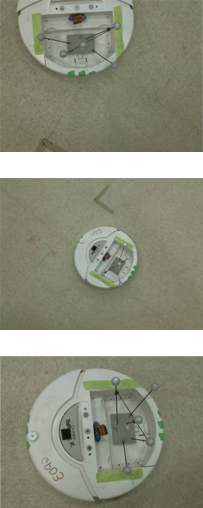}
			\caption{\centering`iRobot' class examples.}
			\label{fig:training_images_overview_v2_irobot}
		\end{subfigure}
		\hfill
		\begin{subfigure}[t]{0.115\textwidth}
			\centering
			\includegraphics[width=1\textwidth]{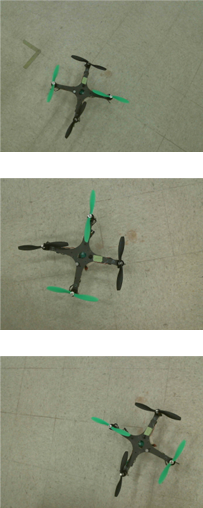}
			\caption{\centering`Quadrotor' class examples.}
			\label{fig:training_images_overview_v2_quad}
		\end{subfigure}
		\hfill
		\begin{subfigure}[t]{0.115\textwidth}
			\centering
			\includegraphics[width=1\textwidth]{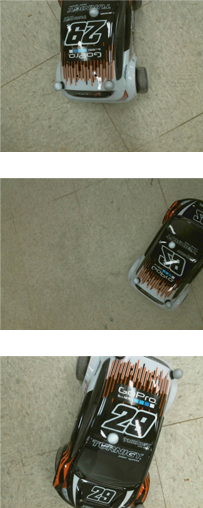}
			\caption{\centering`Racecar' class examples.}
			\label{fig:training_images_overview_v2_racecar}
		\end{subfigure}
		\hfill
		\begin{subfigure}[t]{0.115\textwidth}
			\centering
			\includegraphics[width=1\textwidth]{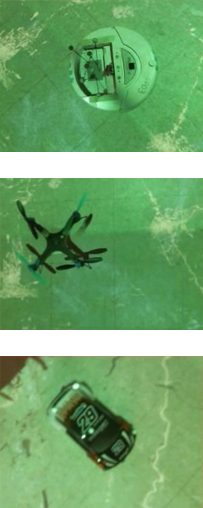}
			\caption{\centering Test dataset examples.}
			\label{fig:training_images_overview_v2_proj_on}
		\end{subfigure}
		\caption[]{First three columns show example training images for classes `iRobot', `Quadrotor', and `Racecar', respectively. Final column shows example test images (captured from a moving quadrotor) composed of the same objects placed in an environment with new lighting conditions.}
		\label{fig:training_images_overview}
	\end{figure}

	Classification robustness is tested by using a projected augmented reality environment to change the lighting conditions of the experiment domain. In comparison to the well-lit images in the training dataset (\cref{fig:training_images_overview_v2_irobot,fig:training_images_overview_v2_quad,fig:training_images_overview_v2_racecar}), the test images have new lighting conditions and textured backgrounds which also tend to reduce the shutter speed of the camera, increasing image blur (\cref{fig:training_images_overview_v2_proj_on}). These experiments are designed to simulate a typical scenario in robotics where the training dataset may not be fully representative of settings encountered during the mission. Robustness against such sources of noise and varying lighting conditions is a fundamental need for robots designed to operate in real-world settings.

	\begin{figure*}[t]
		\centering
		\begin{subfigure}[t]{0.24\textwidth}
			\centering
			\includegraphics[width=1\textwidth]{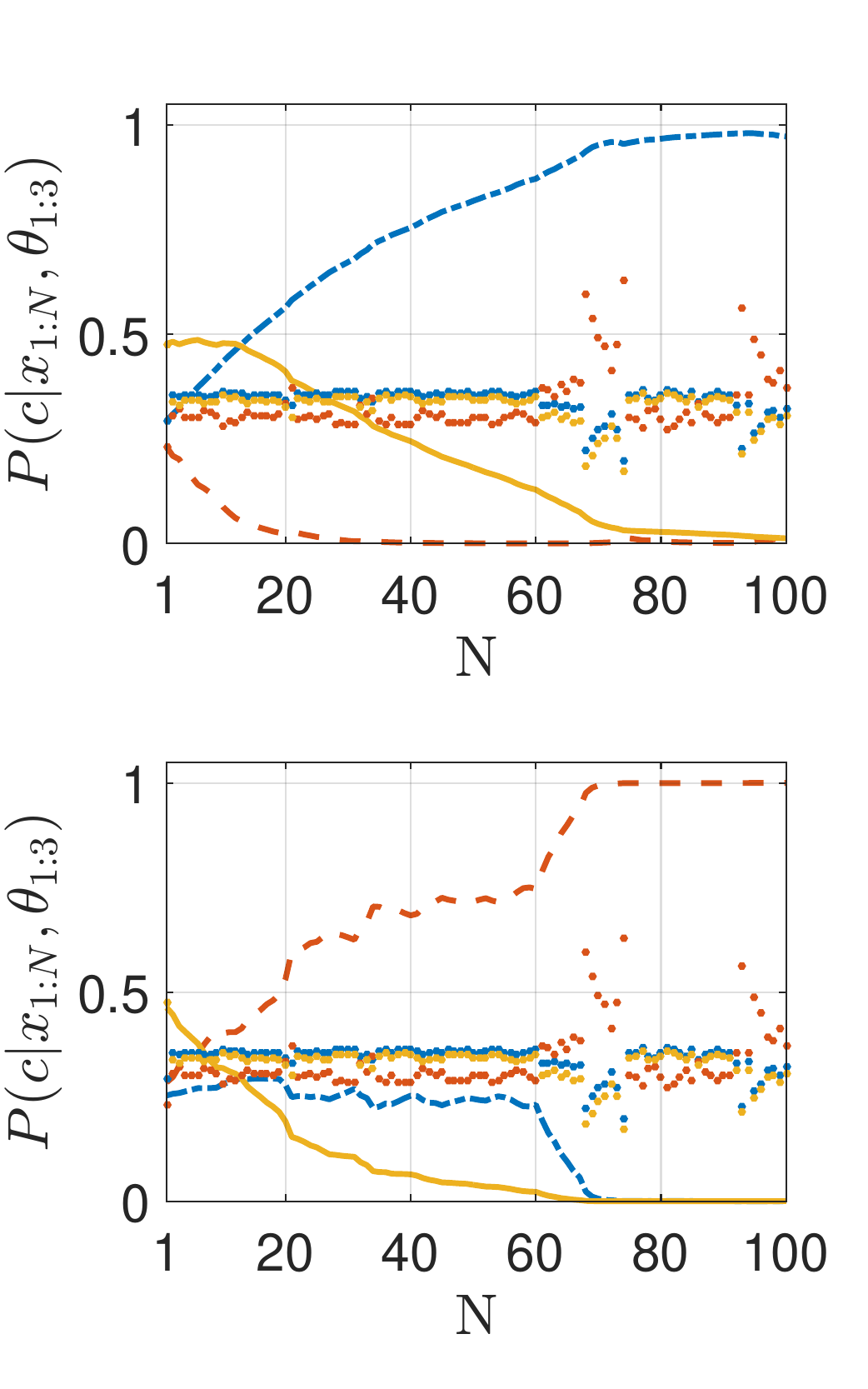}
			\caption{True class is `Quadrotor'.}
			\label{fig:method_comparison_actual_data_true_class_quad}
		\end{subfigure}
		\hfill
		\begin{subfigure}[t]{0.24\textwidth}
			\centering
			\includegraphics[width=1\textwidth]{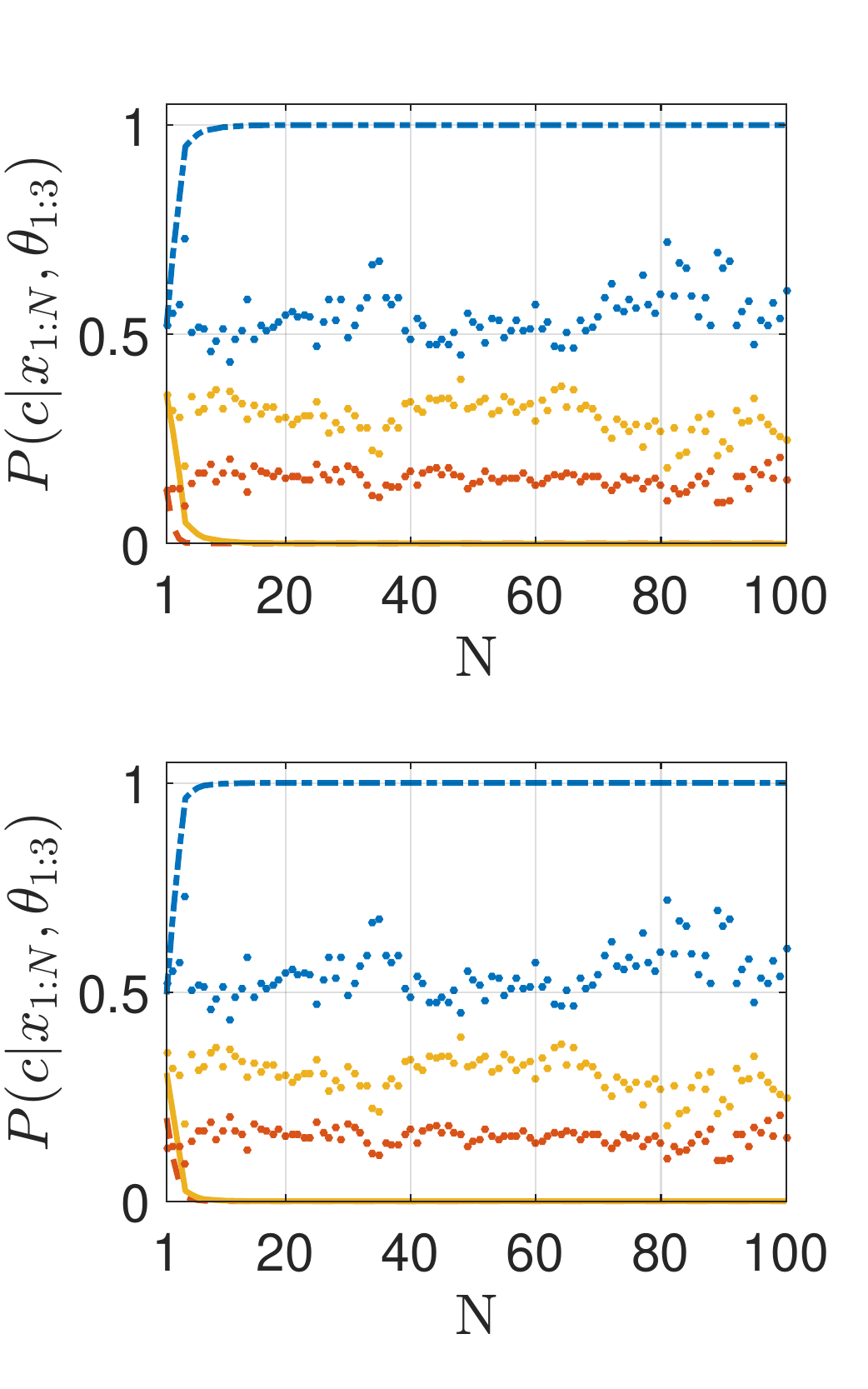}
			\caption{True class is `iRobot'.}
			\label{fig:method_comparison_actual_data_true_class_gpucc}
		\end{subfigure}
		\hfill
		\begin{subfigure}[t]{0.24\textwidth}
			\centering
			\includegraphics[width=1\textwidth]{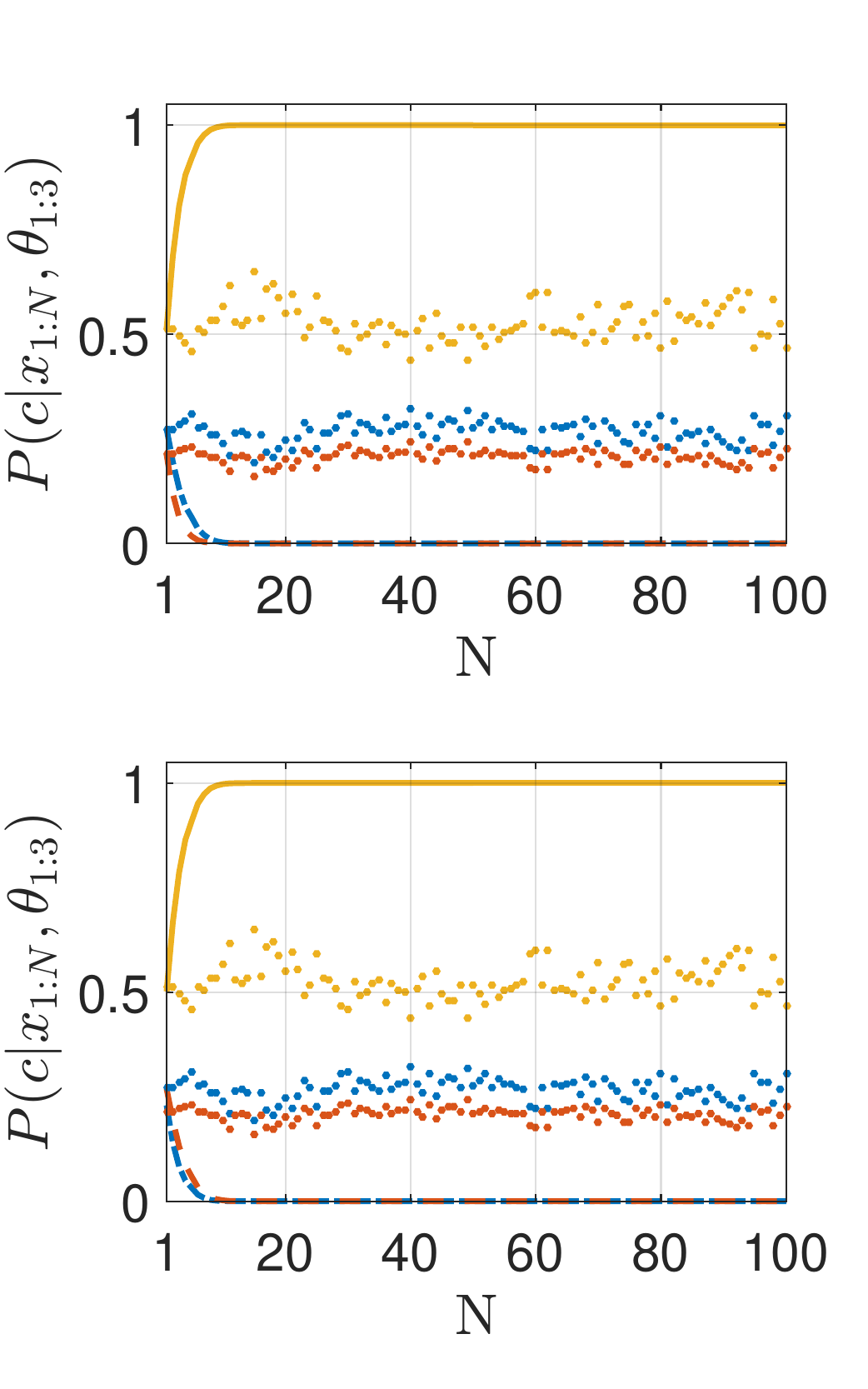}
			\caption{True class is `Racecar'.}
			\label{fig:method_comparison_actual_data_true_class_racecar}
		\end{subfigure}
		\hfill
		\begin{subfigure}[t]{0.24\textwidth}
			\centering
			\includegraphics[width=1\textwidth]{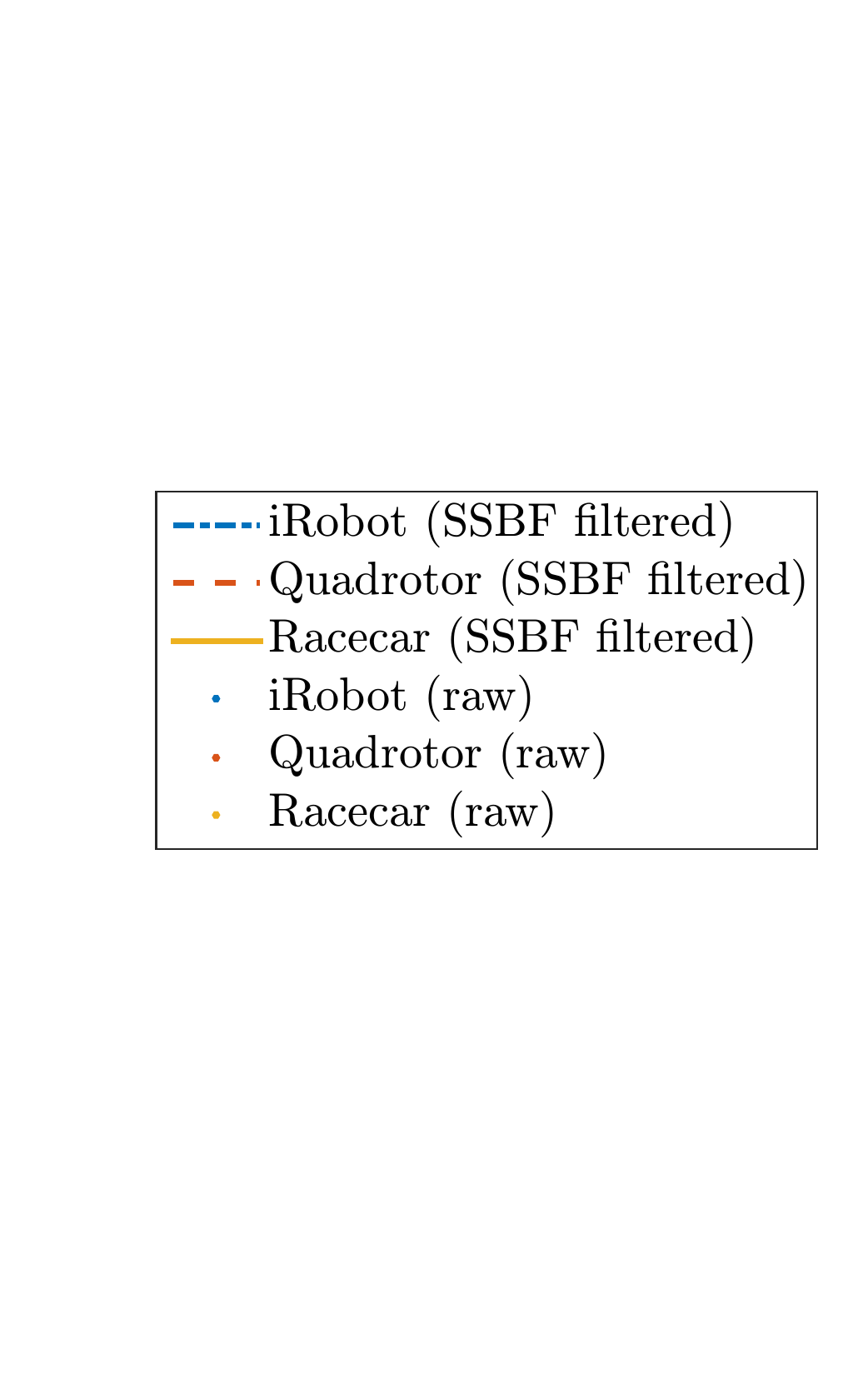}
		\end{subfigure}
		\caption{Classification filtering hardware results from images recorded on a moving quadrotor (figure best viewed in electronic form). Upper and lower plots show SSBF and HBNI filtering, respectively. Points indicate raw classifier outputs $x_i$ and lines represent filtered results. Identical raw classification outputs are used to compare the two methods. The `Quadrotor' object (\cref{fig:method_comparison_actual_data_true_class_quad}) is particularly difficult to classify using the SSBF approach, which misclassifies it as an `iRobot' (\cref{fig:method_comparison_actual_data_true_class_quad}, top). HBNI correctly infers the underlying distribution after only a few observations (\cref{fig:method_comparison_actual_data_true_class_quad}, bottom). Raw classifier outputs $x_i$ for the `iRobot' and `Racecar' classes (\cref{fig:method_comparison_actual_data_true_class_gpucc,fig:method_comparison_actual_data_true_class_racecar}) generally show high confidence in the respective underlying object classes, resulting in almost identical SSBF and HBNI filtering results.}
		\label{fig:method_comparison_hardware}
	\end{figure*}
	
	\begin{figure*}[t]
		\centering
		\includegraphics[width=1\textwidth]{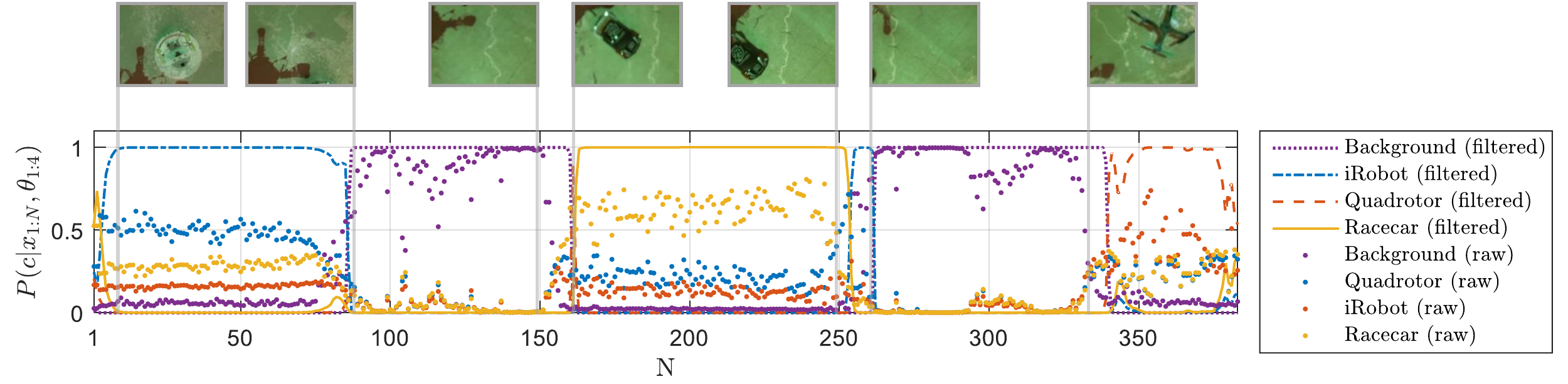}
		\caption{Classification of objects in a scene from onboard a moving quadrotor (example frames indicated). A sliding window approach allows dynamic classification of objects, with latest 12 image frames used to calculated underlying object class.}
		\label{fig:fhandle_rss_python_plots}
	\end{figure*}
	
	Filtered classification results for the test dataset are shown in \cref{fig:method_comparison_hardware}. In these new lighting conditions, classification of the `Quadrotor' object class is particularly difficult, resulting in nearly equal classification probabilities $x_i$ amongst all three classes (raw data in \cref{fig:method_comparison_actual_data_true_class_quad}). A noise-agnostic filter such as SSBF fails at correctly classifying the object as a `Quadrotor', instead classifying it as an `iRobot' with high confidence (filtered output in \cref{fig:method_comparison_actual_data_true_class_quad}, top). Moreover, the probability of the `Quadrotor' class asymptotically approaches zero as more observations are made. In contrast, HBNI allows inference of the underlying noise, leading to robust classification of the `Quadrotor' object class after only 7 frames (\cref{fig:method_comparison_actual_data_true_class_quad}, bottom). In the $N=70$ to $N=75$ range, due to improved lighting, the CNN-based classifier assigns higher probability to the `Quadrotor' class in its raw outputs. SSBF only slightly lowers its probability of the object being an `iRobot', whereas the proposed HBNI approach significantly increases the probability of the true `Quadrotor' class. 
	
	In situations where raw class probabilities $x_i$ show high confidence in a given object class (low noise case), SSBF and HBNI filtered outputs are near-identical (\cref{fig:method_comparison_actual_data_true_class_gpucc,fig:method_comparison_actual_data_true_class_racecar}). This demonstrates that HBNI does not suffer from loss of classification accuracy when test data is near-nominal (i.e., where traditional filtering methods would also succeed).
	
	\cref{fig:fhandle_rss_python_plots} shows HBNI classification results on a quadrotor exploring an environment with multiple objects. A sliding window filtering approach is used to allow detection of new objects in the camera's field of view. In this example, the latest 12 frames are used to infer the underlying object class as the quadrotor explores the domain. The results indicate that HBNI can be used to accurately classify objects onboard a moving robot operating in noisy domains.
	

		
	\section{Conclusion}
	This paper presented Hierarchical Bayesian Noise Inference (HBNI), a method for sequential probabilistic classification. HBNI leverages underlying classifier output noise distributions to increase classification accuracy, even when given a limited number of observations. This is especially useful in robotics, where captured image sequences are prone to noise and training data may not be fully representative of test data. The paper also demonstrated the proposed approach running in real-time on a moving quadrotor platform, with classification and filtering performed onboard at approximately 13 frames per second.	Although HBNI was motivated by the object perception problem in robotics, it can be trivially extended to any other probabilistic classification setting, allowing a wide suite of application domains. Future work includes integration of filtered classification observations into real-time, probabilistic planning frameworks.

	\clearpage
	\bibliographystyle{plainnat}
	\bibliography{references}
\end{document}